\documentclass[sigconf,natbib=false]{acmart}
\AtBeginDocument{%
  }

\usepackage{libertine} 
\usepackage[T1]{fontenc}
\usepackage{textcomp}
\usepackage{amsmath}
\usepackage{array}
\usepackage[caption=false,font=normalsize,labelfont=sf,textfont=sf]{subfig}
\usepackage{textcomp}
\usepackage{stfloats}
\usepackage{url}
\usepackage{verbatim}
\usepackage{graphicx}
\usepackage[normalem]{ulem}
\usepackage{xcolor}
\usepackage{xspace}
\usepackage[ruled,vlined]{algorithm2e}
\usepackage{mathtools}
\usepackage{comment}
\usepackage{tabularx}
\usepackage{tabu}                      
\usepackage{booktabs}                  

\DeclarePairedDelimiter\ceil{\lceil}{\rceil}

\newcommand{\eg}                {\emph{e.g.},\xspace}
\newcommand{\etal}              {\emph{et~al.}\xspace}

\copyrightyear{2025}
\acmYear{2025}
\setcopyright{rightsretained}
\acmConference[GECCO '25 Companion]{Genetic and Evolutionary Computation Conference}{July 14--18, 2025}{Malaga, Spain}
\acmBooktitle{Genetic and Evolutionary Computation Conference (GECCO '25 Companion), July 14--18, 2025, Malaga, Spain}\acmDOI{10.1145/3712255.3726720}
\acmISBN{979-8-4007-1464-1/2025/07}

\RequirePackage[
  datamodel=acmdatamodel,
  style=acmnumeric,
  ]{biblatex}

\begin{document}

\title{Smart Starts: Accelerating Convergence through Uncommon Region Exploration}

\author{Xinyu Zhang}
\orcid{0000-0002-7475-8979}
\affiliation{%
  \institution{Stony Brook University}
  \city{Stony Brook}
  \state{New York}
  \country{USA}
}
\email{zhang146@cs.stonybrook.edu}

\author{M\'ario Antunes†}
\affiliation{%
  \institution{University of Aveiro}
  \city{Aveiro}
  \country{Portugal}}
\email{mario.antunes@ua.pt}

\author{Tyler Estro}
\affiliation{%
  \institution{Stony Brook University}
  \city{Stony Brook}
  \state{New York}
  \country{USA}
}
\email{testro@cs.stonybrook.edu}

\author{Erez Zadok}
\affiliation{%
 \institution{Stony Brook University}
  \city{Stony Brook}
  \state{New York}
  \country{USA}
 }
\email{ezk@cs.stonybrook.edu}

\author{Klaus Mueller}
\affiliation{%
  \institution{Stony Brook University}
  \city{Stony Brook}
  \state{New York}
  \country{USA}
  }
\email{mueller@cs.stonybrook.edu}

\thanks{† Corresponding author.}

\renewcommand{\shortauthors}{Zhang \etal}

\begin{abstract}
Initialization profoundly affects evolutionary algorithm (EA) efficacy by dictating search trajectories and convergence. This study introduces a hybrid initialization strategy combining empty-space search algorithm (ESA) and opposition-based learning (OBL). OBL initially generates a diverse population, subsequently augmented by ESA, which identifies under-explored regions. This synergy enhances population diversity, accelerates convergence, and improves EA performance on complex, high-dimensional optimization problems. Benchmark results demonstrate the proposed method's superiority in solution quality and convergence speed compared to conventional initialization techniques.
\end{abstract}

\ccsdesc[300]{Theory of computation~Theory of randomized search heuristics}
\ccsdesc[300]{Theory of computation~Random search heuristics}

\keywords{Evolutionary algorithms, initialization, opposition-based learning, empty-space search}


\maketitle

\section{INTRODUCTION}

Evolutionary algorithms (EAs) are essential for complex, high-dimensional optimization across diverse domains. Their population-based search, simulating natural evolution, offers robust global exploration, particularly for multimodal and non-convex problems~\cite{beheshti2013review}. However, initial population distribution critically impacts EA performance in intricate landscapes. Concentrated populations risk premature convergence, whereas diverse populations enhance search space coverage~\cite{Agushaka2022}. Effective initialization is thus crucial, especially for rugged, multi-modal fitness landscapes.

Traditional uniform random initialization often fails in high-dimensional spaces. Systematic methods like Latin Hypercube Sampling (LHS)~\cite{ye1998orthogonal} and Sobol sequences~\cite{sobol1967distribution} improve distribution but suffer from the ``curse of dimensionality''. To address this, researchers explore under-explored region identification, including niching methods for multimodal problems~\cite{Shir2012, shir2010adaptive, stoean2007disburdening}. Kazimipour \etal categorized initialization methods by randomness, compositionality, and generality~\cite{6900618}. Recent studies include orthogonal array-based initialization~\cite{KUMAR2022101010} and interval-based multi-objective initialization~\cite{XUE2022109420}.

Compositional methods, like opposition-based learning (OBL)~\cite{MAHDAVI20181}, enhance diversity by generating ``opposite'' solutions. However, OBL may miss under-explored subregions in complex landscapes. The Empty-space Search Algorithm (ESA)~\cite{zhang2025intothevoid}, designed for data visualization, can complement OBL by targeting these regions.

This paper proposes a novel initialization strategy combining OBL and ESA. OBL generates an initial diverse population, subsequently augmented by ESA to fill under-explored areas. This hybrid approach aims to accelerate convergence and enhance EA performance in complex, high-dimensional optimization. Experimental results on benchmark problems, compared with conventional techniques, validate the method's efficacy.

The paper is structured as follows: Section \ref{sec:prelim} reviews relevant technologies; Section \ref{sec:method} details the proposed method; Section \ref{sec:exps} presents experimental results and statistical analysis; and Section \ref{sec:conclusion} summarizes the findings.

\section{PRELIMINARIES}
\label{sec:prelim}
This section lays the groundwork for the proposed methodology by outlining two fundamental concepts: Opposition-Based Learning (OBL) and Empty-space Search Algorithm (ESA) in Sections \ref{sec:obl} \textcolor{blue}{and} \ref{sec:esa} respectively.  

\subsection{Opposition-Based Learning}
\label{sec:obl}


Opposition-Based Learning (OBL)~\cite{MAHDAVI20181} accelerates optimization by simultaneously evaluating candidate solutions $x$ and their opposites $x^{\prime}$. For $x \in R^d$ within a bounded search space $S=[a_1, b_1] \times [a_2, b_2] \times \dots \times [a_d, b_d]$, the opposite $x^{\prime}$ is defined as:
$$
x^{\prime}_i = a_i + b_i - x_i, \forall i \in {1, 2, \dots, d}
$$
where $a_i$ and $b_i$ are the $i-$th dimension's bounds. The superior solution, based on the objective function $f$, is selected, effectively doubling exploration without increasing population size.


\subsection{Empty-space Search Algorithm}
\label{sec:esa}


The Empty-space Search Algorithm (ESA)~\cite{zhang2025intothevoid} is a heuristic method designed to identify sparse, under-explored regions (``empty spaces'') within high-dimensional search spaces. While Zhang \etal applied ESA to identify novel high quality solutions, we find that locating these regions can enhance candidate solution diversity and quality in optimization.


In a data space, ESA first randomly places some points called ``agents'' and then employs a physics-based approach called Lennard-Jones (L-J) Potential to guide the agents to sparse regions. The L-J Potential models the interaction between particles, but here it is used to determine the moving direction of agents. The force of a single data point to an agent is given below:
\begin{equation}
    \label{eq:flj}
    F(r) = 24\frac{1}{\sigma}\left[2\left(\frac{\sigma}{r}\right)^{13} - \left(\frac{\sigma}{r}\right)^7\right]
\end{equation}

where $r$ is the distance of an agent from a data point and $\sigma$ is the particle effect size. The force is positive if $r$ is smaller than $\sigma$, which means that the agent is too close to a data point and will be pushed away, and otherwise the force is negative and pulls the agent back. The final moving direction of the agent is the resultant force direction of $k$ nearest data points:
\begin{equation}
    \label{eq: sumf}
    \begin{aligned}
        \Sigma \Vec{F} &= \sum_{i}^{k}\Vec{u_i}F(r_i) \qquad
        \Vec{d} &= \frac{\Sigma \Vec{F}}{||\Sigma \Vec{F}||}
    \end{aligned}
\end{equation}

where $\Vec{u_i}$ is the unit vector that directs the agent to the data point. Through iterative equilibrium, the agents converge to positions representing the centers of empty spaces.

\section{PROPOSED METHOD}
\label{sec:method}

This section presents OBLESA (OBL+ESA), a novel initialization method designed to enhance EA exploration in complex landscapes. OBLESA addresses the limitation of random or biased initialization by identifying and populating under-explored regions.

Initially, OBL is applied to a random population, generating opposite solutions to expand diversity. Subsequently, ESA, utilizing Lennard-Jones-like potential functions, identifies and populates sparse regions, avoiding dense clusters. The candidate set is temporarily tripled to ensure thorough exploration, then reduced to the top one-third.

OBLESA delivers a higher-quality, diverse initial population, facilitating early exploitation of promising solutions and mitigating premature convergence.

\subsection{ESA for Population Initialization}



Zhang \etal ~\cite{zhang2025intothevoid} designed ESA for Pareto front expansion, employing momentum for trajectory-based sampling. This work adapts ESA to optimize population distribution within empty regions.

Instead of trajectory sampling, we place representative points within each region, avoiding oversampling. We also eliminate momentum, but selecting converged agent positions instead. Furthermore, we replace $k$-nearest neighbor (kNN) with approximate nearest neighbor (ANN) to mitigate high-dimensional query bottlenecks. Algorithm \ref{alg:esa} details this simplified ESA variant.

\begin{algorithm}
\SetAlgoLined

 Set the number of neighbors $k$, the particle effective diameter $\sigma$, the number of search steps $n$, the
 step size $\alpha$, the vanishing threshold $\delta$\;

 Initialize an agent $\pi = \mathbf{c}$
 where $\mathbf{c}$ is a random coordinate\;
 Specify constraints on the target function:\\
 \hspace*{0.5em} $f_1(\pi) <= 0$; $f_2(\pi) <= 0$; \dots, $f_p(\pi) <= 0$\;
 \For {i \dots $n$} {
    
    Get $k$ approximated nearest neighbors of the agent from the dataset\;
    Use Eq. \ref{eq: sumf} to calculate $\Vec{d}$\;
    \If {$||\Sigma \Vec{F}|| < \delta$} {
        return $\tau$\;
    } 

        $\pi = \pi + \Vec{d} * \alpha$\;
    
    \If{$\pi$ violates any constraint $f$} {
        return $\pi$\;
    }
 }
 return $\pi$\;
 \caption{Empty-Space Search for a Single Agent}
 \label{alg:esa}
\end{algorithm}

There are three key parameters in this algorithm that affect the agent convergence: the number of neighbors $k$, the particle effective diameter $\sigma$, and the step size $\alpha$. Zhang \etal recommended setting $k=d+1$ where $d$ is the dimensionality of the problem, $\sigma = $ \textit{average distance to the neighbors}, and $\alpha = 0.01$ in their work. In our experiments, we generally adopted their recommendations, but empirically reduced $\sigma$ to half of average distance to the neighbors to prevent agents from moving out of the bounding box.



\subsection{Initialization Strategy}

High-dimensional empty region counts increase exponentially, a consequence of the \emph{curse of dimensionality}~\cite{Altman2018}. Delaunay Triangulation (DT) reveals a space complexity of 
$O(n^{\ceil{{d/2}}})$ \cite{seidel1995upper}, demonstrating the significant complexity growth with dimension and data size. Thus, thorough space exploration is crucial. Given a population size of $n_{pop}$, we summarize our algorithm in four steps: i) Random population initialization ($n_{pop}$); ii) Opposition point generation; iii) ESA agent deployment and empty-space population augmentation; and iv) Population refinement to $n_{pop}$.

Random initialization, followed by OBL, ensures broad initial coverage. ESA then refines exploration, targeting under-explored regions. 
The population, tripled post-ESA, is evaluated, and the top one-third is retained for subsequent optimization.

\section{EVALUATION}
\label{sec:exps}
 
This section presents a comprehensive evaluation of the proposed methodology. We first detail the experimental setup (\autoref{sec:settings}), including the benchmark functions, parameter settings, and evaluation metrics employed.  Subsequently, we analyze the results obtained (\autoref{sec:results}), comparing the performance of the proposed approach against established baseline algorithms and discussing the key observations and insights gained from the experiments.

\subsection{Evaluation Settings}
\label{sec:settings}

In this section, we present the evaluation results of OBLESA against two baselines: random initialization and random + OBL initialization, using 24 benchmark functions. Specifically, we first selected target functions from COCO benchmark~\cite{hansen2021coco} and generated the initial population using each method, setting the initial population size to $n_{pop}=100$. The generated initial population was then used as input for two evolutionary algorithms: Differential Evolution (DE)~\cite{storn1997differential} and Enhanced Grey Wolf Optimization (EGWO)~\cite{Joshi2017, sharma2022comprehensive} and tested their performance. 

When evaluating initialization strategies, employing both DE and EGWO offers a robust approach due to their distinct characteristics. DE, a well-established evolutionary algorithm, provides a solid baseline, making it a reliable tool for assessing fundamental performance. Conversely, EGWO, representing a state-of-the-art metaheuristic, allows for the examination of how advanced optimization techniques respond to varying initial conditions.

Each function in COCO benchmark has an internal convergence threshold, meaning that the optimization process stops as soon as the threshold is reached. Additionally, each function has multiple variants (\eg rotations) and default budget. 
The optimization process terminates immediately either upon reaching the convergence threshold or when the budget is exhausted. 
To ensure consistency, we left all default settings in the COCO benchmark, including the internal convergence threshold, budget, and problem dimensionality (2D, 3D, 5D, 10D, 20D, and 40D). We measured performance by calculating the fraction of benchmark functions that reached the convergence threshold during the optimization process.

To account for the randomness inherent in evolutionary algorithms, we ran each initialization strategy 10 times with different random seeds (1 - 10). 

\subsection{Result Analysis}
\label{sec:results}

We put our results in the supplementary material \footnote{The supplementary material is available at \url{github.com/Lagrant/SmartStart}.} where we observe that the performance of both algorithms degrades as dimensionality increases. However, EGWO consistently outperforms DE across all dimensions. In addition, OBLESA initialization is better than the other two baselines in most cases for both EGWO and DE across the six dimensionalities. 

We also notice that the advantage lead by OBLESA is mostly small. To assess the impact of randomness, we aggregated the results of 10 seeds and performned a statistical analysis to test the significance of the difference between OBLESA and the baselines. Specifically, we assigned scores to each initialization strategy based on their rank per dimensionality per seed. For instance, if OBLESA ranked first and Random ranked third, OBLESA received 3 points while Random received 1. Next we summed the scores over 10 seeds for each dimensionality. Then we did ANOVA and post-hoc test to analyze whether the differences are statistically significant. The results are listed in Tab. \ref{tab:egwo} and Tab. \ref{tab:de}. 

Overall, OBLESA generally achieved higher scores than the baselines. The ANOVA and the associated post-hoc test show that OBLESA is similar to baselines, but it leads significant advantage in higher dimensional space. As expected given the nature of ESA, it should provide greater benefits the larger the problem dimensionality.

\begin{table*}[tb]
  \caption{Advantage score of each initialization strategy on EGWO. The p-value shows the statistical significance of the difference calculated by ANOVA. OBLESA-OBL and OBLESA-RANDOM are the post-hoc analysis results. Each entry in the second table is the p-value to analyze the significance of pairwise difference.}
  \label{tab:egwo}
  \scriptsize%
  \centering%
  \begin{tabu} to \textwidth {X[c] *{6}{X[c]}}
  	\toprule
  	Strategy &  2D	 &   3D	 &   5D & 10D & 20D & 40D  \\
  	\midrule
  	OBLESA & 21 & 23 & 17	 &  24	 & 27 & 29 \\
        OBL & 20 & 20 & 23 & 17 & 21 & 16 \\
        RANDOM & 19 & 17 & 20 & 19 & 12 & 15 \\
        p-vaule & 0.873 & 0.280 & 0.280 & 0.153 & $<0.0001$ & $<0.0001$ \\
  	\bottomrule
   
  \end{tabu}%

    \begin{tabu} to \textwidth {X[c] *{6}{X[c]}}
  	\toprule
  	post-hoc &  2D	 &   3D	 &   5D & 10D & 20D & 40D  \\
  	\midrule
  	OBLESA -OBL & 0.963 & 0.696 & 0.249	 &  0.144	 & 0.062 & $<0.0001$ \\
        OBLESA -RANDOM & 0.861 & 0.249 & 0.696 & 0.359 & $<0.0001$ & $<0.0001$ \\
  	\bottomrule
   
  \end{tabu}%
  
\end{table*}

\begin{table*}[tb]
  \caption{Advantage score of each initialization strategy on DE. The p-value shows the statistical significance of the difference calculated by ANOVA. OBLESA-OBL and OBLESA-RANDOM are the post-hoc analysis results. Each entry in the second table is the p-value to analyze the significance of pairwise difference.}
  \label{tab:de}
  \scriptsize%
  \centering%
  \begin{tabu} to \textwidth {X[c] *{6}{X[c]}}
  	\toprule
  	Strategy &  2D	 &   3D	 &   5D & 10D & 20D & 40D  \\
  	\midrule
  	OBLESA & 19 & 23 & 26	 &  28	 & 23 & 29 \\
        OBL & 22 & 21 & 18 & 18 & 19 & 18 \\
        RANDOM & 19 & 16 & 16 & 14 & 18 & 13 \\
        p-vaule & 0.663 & 0.153 & 0.119 & $<0.0001$ & 0.375 & $<0.0001$ \\
  	\bottomrule
   
  \end{tabu}%

      \begin{tabu} to \textwidth {X[c] *{6}{X[c]}}
  	\toprule
  	post-hoc &  2D	 &   3D	 &   5D & 10D & 20D & 40D  \\
  	\midrule
  	OBLESA -OBL & 0.711 & 0.844	 &  0.053	 & 0.002 & 0.536 & $<0.0001$ \\
        OBLESA -RANDOM & 1 & 0.144 & 0.013 & $<0.0001$ & 0.382 & $<0.0001$ \\
  	\bottomrule

  \end{tabu}%
\end{table*}

In our experiments, we also find that OBLESA does not show a noticeable advantage against the baselines. Some possible factors could be (1) the refinement step: we simply evaluated the entire augmented population and chose the top one third after OBL and ESA. It may lead to some mode collapse since the landscape of a target function in high-dimensional space is complex. Simply choosing the best data probably traps the algorithm to local optima. In the follow-up research, we will try different ways to refine the population, \eg assign a probability to reject high-fitness data points or pair a random point with its associated opposite point and empty-space agent, choose the best one among the three candidates in each pair; (2) the initial population size: in our experiments, we simply set $n_{pop}=100$ to save time and computational resources. However, due to counterintuitive facts in high-dimensional space \cite{anderson2024counterintuitive, verleysen2005curse}, the empty space will be much more complex. Increasing the population size can probably improve performance; (3) parameter fine-tuning: we adopted most parameters recommended by Zhang \etal in their work, but they aim to find Pareto optimal configurations. The goal differs from our work. Therefore, further investigation is necessary about the parameter setting.

In further research, we will work more on the aspects mentioned above and look deeply into the algorithm mechanism to improve OBLESA's performance across the benchmarks. The algorithm implementation and experiment settings are also available in the supplementary material.

\section{CONCLUSION}
\label{sec:conclusion}

The experimental results highlight the effectiveness of OBLESA in improving population diversity and accelerating convergence in evolutionary algorithms. By integrating Opposition-Based Learning (OBL) with the Empty-Space Search Algorithm (ESA), OBLESA addresses key limitations of existing initialization strategies in high-dimensional problems. Traditional random sampling ensures broad coverage of the search space but often leaves large unexplored regions, while OBL enhances exploration by generating complementary solutions yet lacks an explicit mechanism to target under-sampled areas. ESA bridges this gap by actively identifying and filling the sparse regions, leading to a more balanced and representative initial population.

Our results demonstrate that OBLESA significantly improves optimization performance, particularly in high-dimensional spaces (20D and 40D). Statistical analysis confirms that these improvements are not incidental, as OBLESA consistently outperforms baselines in these challenging settings. However, in lower-dimensional problems, the advantage of OBLESA is less pronounced, suggesting that its benefits become increasingly relevant as the complexity of the search space grows.





\begin{acks}
This work was supported by FCT - Fundação para a Ciência e Tecnologia, I.P. by project reference UIDB/50008, and DOI identifier 10.54499/UIDB/50008.
\end{acks}

\printbibliography

\end{document}